\documentstyle[eaclap]{article}

\title{\vspace{-0.5in}Higher-order Linear Logic Programming\\
of Categorial
Deduction}
\author{Glyn Morrill\\
Secci\'{o} d'Intel$\cdot$lig\`{e}ncia Artificial\\
Departament de Llenguatges i Sistemes Inform\`{a}tics\\
Universitat Polit\`{e}cnica de Catalunya\\
Pau Gargallo, 5\\
08028 Barcelona\\
morrill@lsi.upc.es\\}

\newenvironment{disp}
   {\begin{equation}\begin{minipage}[t]{2.65in}}%
   {\end{minipage}\end{equation}\vspace{0in}}
\newenvironment{plaindisp}
   {\begin{minipage}[t]{3.75in}}%
   {\end{minipage}}
   {\end{minipage}\end{equation}\vspace{0in}\end{footnotesize}}
\newcommand{\verbquote}[1]{``#1''}
\newcommand{\scare}[1]{``#1''}

\newcommand{\prevex}{\arabic{equation}}
\newcommand{\nextex}{\addtocounter{equation}{1}\arabic{equation}\addtocounter{equation}{-1}}

\newcommand{\vsl}{\mbox{\mbox{$|$}}}
\newcommand{\bsl }{\mbox{\mbox{$\backslash$}}}

\newcommand{\tab}{\mbox{\hspace*{0.35in}}}
\newcommand{\tb}{\mbox{\hspace*{0.25in}}}

\newcommand{\boolpicklen}[4]
   {\setlength{#1}{#4}
    \setlength{#1}{-1#1}
    \addtolength{#1}{#3}
    \setlength{#1}{#2#1}
    \addtolength{#1}{#4}}

\newcounter{nolabelbool}

\newlength{\exwidth}

\newcommand{\exonecolumn}%
   {\setlength{\exwidth}{\textwidth}
    \addtolength{\exwidth}{-0.5in}}

\newcommand{\extwocolumn}%
   {\setlength{\exwidth}{0.5\textwidth}
    \addtolength{\exwidth}{-0.5\columnsep}
    \addtolength{\exwidth}{-0.4in}}

\exonecolumn

   {\end{list}
    \end{minipage}
    \end{equation}
    \setcounter{nolabelbool}{0}}

   {\end{list}
    \end{minipage}
    \setcounter{nolabelbool}{0}
    \vspace*{.18in}\\}

   {\end{list}
    \end{footnotesize}
    \end{minipage}
    \end{equation}
    \setcounter{nolabelbool}{0}}

\newlength{\lenabc}

\newcommand{\lingform}[1]{`#1'}

\newlength{\dtopskip}
\newlength{\dbottomskip}
\newlength{\dheight}
\newlength{\lena}

\newlength{\firstlen}
\newlength{\secondlen}
\newlength{\dmargin}
\newlength{\dmarginbasic}
\newlength{\wordgap}
\newlength{\spacing}
\newlength{\dlineskip}
\newlength{\textlen}
\newlength{\vertpos}
\newlength{\centrept}
\newlength{\dlinelen}
\newlength{\catlen}
\newlength{\dlineraise}
\newlength{\dlinethickness}
\newlength{\lone}
\newlength{\ltwo}
\newlength{\lthree}
\newlength{\rone}
\newlength{\rtwo}
\newlength{\rthree}
\newlength{\lfour}
\newlength{\rfour}
\newlength{\lfive}
\newlength{\rfive}
\newlength{\lsix}
\newlength{\rsix}
\newlength{\lseven}
\newlength{\rseven}
\newlength{\leight}
\newlength{\reight}
\newlength{\lnine}
\newlength{\rnine}
\newlength{\lten}
\newlength{\rten}

\newcommand{\settolargerlength}[3]{
   \setlength{\firstlen}{#2}
   \addtolength{\firstlen}{-1#3}
   \setlength{\secondlen}{#3}
   \addtolength{\secondlen}{-1#2}
   \settowidth{\firstlen}{\framebox[\firstlen]{}\framebox[\secondlen]{}}
   \addtolength{\firstlen}{#2}
   \addtolength{\firstlen}{#3}
   \setlength{#1}{0.5\firstlen}}

\newcommand{\putdobj}[4]{
    \setlength{\vertpos}{\dtopskip}
    \addtolength{\vertpos}{#1\dlineskip}
    \addtolength{\vertpos}{#2}
    \put(0,0){\raisebox{\vertpos}{\rule{#3}{0in}{#4}}}}

\newcommand{\putdobjcentred}[4]{
   \putdobj{#1}{#2}{#3}{\makebox[0in]{#4}}}

\newcommand{\putdparbox}[5]{
   \putdobj{#1}{#2}{#3}{\parbox[t]{#4}{#5}}}

\newcommand{\derivline}[1]{\rule[\dlineraise]{#1}{\dlinethickness}}

\newenvironment{deriv}[1]%
  {\noindent
   \begin{minipage}[t]{1in}
   \setlength{\wordgap}{1.5em}%
   \setlength{\dlineraise}{0.265em}
   \setlength{\dlineskip}{-0.9em}
   \setlength{\dtopskip}{\dlineskip}
   \setlength{\dbottomskip}{0\dlineskip}
   \setlength{\dlinethickness}{0.01in}
   \setlength{\dmarginbasic}{0in}
   \resetdmargin
   \setlength{\dheight}{-#1\dlineskip}
   \begin{picture}(0,0)}%
  {\end{picture}
   \newline
   \addtolength{\dheight}{-\dtopskip}
   \addtolength{\dheight}{-\dbottomskip}
   \rule{0in}{\dheight}
   \newline
   \end{minipage}
   \newline}

\newenvironment{derivfns}[1]%
  {\begin{footnotesize}
   \begin{deriv}{#1}}%
  {\end{deriv}
   \end{footnotesize}}

\newcommand{\shiftdmargin}[1]{
   \setlength{\lena}{#1}
   \addtolength{\dmargin}{\lena}}

\newcommand{\resetdmargin}{\setlength{\dmargin}{\dmarginbasic}}

\newcommand{\preword}[1]{
  \putdobj{0}{0in}{\dmargin}{#1}
  \settowidth{\textlen}{#1}
  \shiftdmargin{\textlen}
  \shiftdmargin{\wordgap}}

\newcommand{\derivitem}[1]{\preword{#1}}

\newcommand{\word}[5]{
  \settowidth{\textlen}{#3}
  \settowidth{\catlen}{#4}
  \settowidth{\spacing}{#5}
  \settolargerlength{\dlinelen}{\textlen}{\catlen}
  \settolargerlength{\spacing}{\spacing}{\dlinelen}
  \setlength{\centrept}{\dmargin}
  \addtolength{\centrept}{0.5\spacing}
  \putdobjcentred{0}{0in}{\centrept}{#3}
  \putdobjcentred{1}{0in}{\centrept}{\derivline{\dlinelen}}
  \putdobjcentred{2}{0in}{\centrept}{#4}
  \setlength{#1}{\centrept}
  \addtolength{#1}{-0.5\catlen}
  \setlength{#2}{\centrept}
  \addtolength{#2}{0.5\catlen}
  \shiftdmargin{\spacing}
  \shiftdmargin{\wordgap}}

\newcommand{\wordone}[3]{\word{\lone}{\rone}{#1}{#2}{#3}}
\newcommand{\wordtwo}[3]{\word{\ltwo}{\rtwo}{#1}{#2}{#3}}
\newcommand{\wordthree}[3]{\word{\lthree}{\rthree}{#1}{#2}{#3}}

\newcommand{\cat}[4]{
  \settowidth{\catlen}{#3}
  \settowidth{\spacing}{#4}
  \settolargerlength{\spacing}{\spacing}{\catlen}
  \setlength{\centrept}{\dmargin}
  \addtolength{\centrept}{0.5\spacing}
  \putdobjcentred{0}{0in}{\centrept}{#3}
  \setlength{#1}{\centrept}
  \addtolength{#1}{-0.5\catlen}
  \setlength{#2}{\centrept}
  \addtolength{#2}{0.5\catlen}
  \shiftdmargin{\spacing}
  \shiftdmargin{\wordgap}}

\newcommand{\catone}[2]{\cat{\lone}{\rone}{#1}{#2}}
\newcommand{\cattwo}[2]{\cat{\ltwo}{\rtwo}{#1}{#2}}
\newcommand{\catthree}[2]{\cat{\lthree}{\rthree}{#1}{#2}}
\newcommand{\catfour}[2]{\cat{\lfour}{\rfour}{#1}{#2}}
\newcommand{\catfive}[2]{\cat{\lfive}{\rfive}{#1}{#2}}
\newcommand{\catsix}[2]{\cat{\lsix}{\rsix}{#1}{#2}}

\newcommand{\reduce}[5]{
  \setlength{\dlinelen}{#3}
  \addtolength{\dlinelen}{-1#2}
  \settowidth{\catlen}{#5}
  \settolargerlength{\dlinelen}{\dlinelen}{\catlen}
  \setlength{\centrept}{0.5#2}
  \addtolength{\centrept}{0.5#3}

\putdobjcentred{#1}{0in}{\centrept}{\derivline{\dlinelen}\makebox[0in][l]{#4}}
  \putdobjcentred{#1}{\dlineskip}{\centrept}{#5}
  \setlength{#2}{\centrept}
  \addtolength{#2}{-0.5\catlen}
  \setlength{#3}{\centrept}
  \addtolength{#3}{0.5\catlen}}

\renewcommand{\makelabel}[1]%
    {\makebox[0in]{\begin{picture}(0,0)%
                   \put(700,0){\parbox{1in}{\begin{equation}
                                            \label{#1}
					    \end{equation}}}%
                   \end{picture}}}

\newcommand{\putderivnumkey}[3]%
    {\putdparbox{#1}{0.15in}{#2}{3.00in}%
                {\begin{equation}
                 \label{#3}
                 \end{equation}}}

\newcommand{\putderivnum}[2]%
    {\putdparbox{#1}{0.15in}{#2}{3.00in}%
                {\begin{equation}
                 \end{equation}}}

\newcommand{\derivnumkey}[1]{\putderivnumkey{0}{0in}{#1}}

\newcommand{\derivnum}{\putderivnum{0}{0in}}

\newcommand{\axiom}[5]{
  \settowidth{\catlen}{#3}
  \settowidth{\spacing}{#4}
  \settolargerlength{\spacing}{\spacing}{\catlen}
  \setlength{\centrept}{\dmargin}
  \addtolength{\centrept}{0.5\spacing}
  \putdobjcentred{#5}{0in}{\centrept}{#3}
  \setlength{#1}{\centrept}
  \addtolength{#1}{-0.5\catlen}
  \setlength{#2}{\centrept}
  \addtolength{#2}{0.5\catlen}
  \shiftdmargin{\spacing}
  \shiftdmargin{\wordgap}}

\newcommand{\axiomone}[3]{\axiom{\lone}{\rone}{#2}{#3}{#1}}
\newcommand{\axiomtwo}[3]{\axiom{\ltwo}{\rtwo}{#2}{#3}{#1}}
\newcommand{\axiomthree}[3]{\axiom{\lthree}{\rthree}{#2}{#3}{#1}}
\newcommand{\axiomfour}[3]{\axiom{\lfour}{\rfour}{#2}{#3}{#1}}
\newcommand{\axiomfive}[3]{\axiom{\lfive}{\rfive}{#2}{#3}{#1}}
\newcommand{\axiomsix}[3]{\axiom{\lsix}{\rsix}{#2}{#3}{#1}}
\newcommand{\axiomseven}[3]{\axiom{\lseven}{\rseven}{#2}{#3}{#1}}
\newcommand{\axiomeight}[3]{\axiom{\leight}{\reight}{#2}{#3}{#1}}
\newcommand{\axiomnine}[3]{\axiom{\lnine}{\rnine}{#2}{#3}{#1}}

\newcommand{\xmod}[1]%
     {\mbox{${\setlength{\fboxsep}{0.5mm} \:
      \fbox{{\tiny \raisebox{.2em}{#1}}\rule[.23ex]{0ex}{0.8ex}} \:}$}}

\newcommand{\nolinereduce}[4]{
  \settowidth{\catlen}{#4}
  \setlength{\centrept}{0.5#2}
  \addtolength{\centrept}{0.5#3}
  \putdobjcentred{#1}{0in}{\centrept}{#4}
  \setlength{#2}{\centrept}
  \addtolength{#2}{-0.5\catlen}
  \setlength{#3}{\centrept}
  \addtolength{#3}{0.5\catlen}}

\newcommand{\yields}{\mbox{$\ \Rightarrow\ $}}

\newcommand{\lnk}{\mbox{\ --\ }}
\newcommand{\ass}{\mbox{:\ }}

\newcommand{\product}{\mbox{$\bullet$}}

\newcommand{\nao}{\mbox{$<$}}
\newcommand{\nau}{\mbox{$>$}}

\newcommand{\naprod}{\mbox{$\stackrel{\diamond}{}$}}

\newcommand{\wrap}{\mbox{$\uparrow$}}
\newcommand{\infix}{\mbox{$\downarrow$}}
\newcommand{\dprod}{\mbox{$\odot$}}

\newcommand{\mybrack}[1]{\mbox{[$\,$#1$\,$]}}
\newcommand{\antibrack}[1]{\mbox{[$\,$#1$\,$]$^{-1}$}}

\newcommand{\nil}{\mbox{$\epsilon$}}
\newcommand{\exsep}{\vspace{-0.125in}}

\newcommand{\NL}{\mbox{\bf NL}}
\newcommand{\AL}{\mbox{\bf L}}
\newcommand{\add}{\mbox{\small $+$}}
\newcommand{\mconj}{\mbox{$\,\otimes\,$}}

\newcommand{\ent}[1]{\begin{minipage}{3in}
\hspace*{0.125in}#1\end{minipage}}

{\end{minipage}\vspace{0in}\newline}

\newcommand{\lingate}{\mbox{$\,\circ\!-\,$}}
\newcommand{\pcnst}[1]{{\bf #1}}
\newcommand{\pconst}[1]{{\bf #1}}
\newcommand{\lnksm}{\mbox{-}}

\begin{document}

\maketitle
\vspace{-0.5in}
\begin{abstract}

We show how categorial deduction can be implemented
in higher-order (linear) logic programming, thereby realising
parsing as deduction for
the associative and non-associative Lambek calculi.
This
provides a method of solution to the parsing problem of
Lambek categorial grammar applicable to a variety of its extensions.\\

\end{abstract}

The present work deals with the parsing problem for
Lambek calculus and its extensions as developed in,
for example,
Moortgat (1988),
van Benthem (1991),
Moortgat and Morrill (1991),
Moortgat and Oehrle (1993),
Morrill (1994b) and
Hepple (1995). Some previous approaches to parsing Lambek
grammar such
as K\"{o}nig (1989), Hepple (1990) and Hendriks
(1993) have
concentrated on the possibilities of sequent proof
normalisation. In Roorda (1991), Moortgat (1992),
Hendriks (1993)
and Oehrle (1994) a strategy of unfolding and labelling for proof
net construction is considered. We aim to show here how
such unfolding allows compilation into programs executable
by a version of SLD resolution, implementing categorial deduction
in dynamic linear clauses. The linearity resides
in the use exactly once per word token of each of the clauses compiled
from lexical categorisations.
By dynamic, it is meant
that clauses may be higher-order (they are hereditary Harrop Horn
clauses) so that
clausal resolution involves insertion in, as well as
retraction from, the resolution database; see Miller et al.\ (1991), and Hodas
and Miller (1994).

It is shown how a range of calculi
can be treated by dealing with the highest common factor
of connectives as linear logical validity. The prosodic
(i.e.\ sublinear) aspects of word order and hierarchical
structure are encoded in labels, in effect the term structure
of quantified linear logic. Compiling labels according
to interpretations in groupoids provides a general method
for calculi with various structural properties and also
for multimodal hybrid formulations. Unification must be
carried out according to the structural axioms but is
limited to {\em one-way} matching, i.e.\ one term is
always ground.
Furthermore, for the particular case of associative Lambek
calculus an additional perspective of binary relational
interpretation allows an especially efficient
coding in which
the span of expressions is
represented in such a way as to avoid the computation
of unifiers under associativity, and this can also
be exploited for non-associative calculus.

Higher-order linear logic programming
has already been applied to natural language processing
in, for example, Hodas (1992) and Hodas and
Miller (1994), in work deriving from
Pareschi (1989) and
Pareschi and Miller (1990). What we show here is that such
implementation can be realised systematically, indeed
by a mechanical compilation, while grammars themselves
are written in higher level categorial grammar formalism.

Automated deduction for Lambek calculi is of interest
in its own right but solution of the parsing problem
for categorial logic allowing significant linguistic
coverage demands automated deduction for more than just
individual calculi. There is a need for methods applying
to whole classes of systems in ways which are principled
and powerful enough to support the further generalisations
that grammar development will demand. We aim to indicate
here how higher-order logic programming can provide for such
a need.

After reviewing the \scare{standard} approach, via sequent
proof normalisation, we outline the relevant features
of (linear) logic programming and explain compilation
and execution
for associative and non-associative calculi in terms
of
groupoid and binary relational interpretations of categorial
connectives. We go on to briefly mention multimodal
calculi for the binary connectives.


The parsing problem is usually construed as the recovery
of structural descriptions assigned to strings by
a grammar. In practice the interest is in computing
semantic forms implicit in the structural descriptions,
which are themselves usually implicit in the history of a derivation
recognising well-formedness of a string. This is true
in particular of compositional categorial architectures
and we shall focus on algorithms for showing well-formedness.
The further step to computing semantics is unproblematic.


For the non-associative Lambek calculus {\bf NL} of Lambek
(1961) we assume types freely generated from a set
of primitive types by binary (infix) operators
\bsl, /{} and \product. A sequent comprises a succedent type
$A$ and an antecedent configuration $\Gamma$ which is a binary bracketed
list of one or more types; we write $\Gamma\yields A$. The
notation $\Gamma(\Delta)$ here refers to a configuration
$\Gamma$ with a distinguished subconfiguration
$\Delta$. \\
\begin{derivfns}{2}
\derivnum
\derivitem{a.}
\catone{$A \yields A$}{}
\derivitem{id}
\catone{$\Gamma \yields A$}{}
\cattwo{$\Delta(A) \yields B$}{}
\reduce{1}{\lone}{\rtwo}{Cut}{$\Delta(\Gamma) \yields
B$}
\end{derivfns}
\begin{derivfns}{2}
\derivitem{b.}
\catone{$\Gamma \yields A$}{}
\cattwo{$\Delta(B) \yields C$}{}
\reduce{1}{\lone}{\rtwo}{\bsl L}{$\Delta([\Gamma, A\bsl B])
\yields C$}
\derivitem{}
\catone{$[A, \Gamma] \yields B$}{}
\reduce{1}{\lone}{\rone}{\bsl R}{$\Gamma \yields A\bsl B$}
\end{derivfns}
\begin{derivfns}{2}
\derivitem{c.}
\catone{$\Gamma \yields A$}{}
\cattwo{$\Delta(B) \yields C$}{}
\reduce{1}{\lone}{\rtwo}{/L}{$\Delta([B/A, \Gamma]) \yields C$}
\derivitem{}
\catone{$[\Gamma, A] \yields B$}{}
\reduce{1}{\lone}{\rone}{/R}{$\Gamma \yields B/A$}
\end{derivfns}
\begin{derivfns}{2}
\derivitem{d.}
\catone{$\Gamma([A, B]) \yields C$}{}
\reduce{1}{\lone}{\rone}{$\product$L}{$\Gamma(A\product B) \yields C$}
\derivitem{}
\catone{$\Gamma\ \yields A$}{}
\cattwo{$\Delta \yields B$}{}
\reduce{1}{\lone}{\rtwo}{$\product$R}{$\Gamma,\ \Delta \yields A \product B$}
\end{derivfns}
For the associative Lambek calculus {\bf L} of
Lambek (1958) the types are the same. A sequent comprises a succedent type $A$
and an antecedent configuration $\Gamma$ which is a list of one or more
types; again we write $\Gamma\yields A$.\\
\begin{derivfns}{2}
\derivnumkey{AL}
\derivitem{a.}
\catone{$A \yields A$}{}
\derivitem{id}
\catone{$\Gamma \yields A$}{}
\cattwo{$\Delta(A) \yields B$}{}
\reduce{1}{\lone}{\rtwo}{Cut}{$\Delta(\Gamma) \yields
B$}
\end{derivfns}
\begin{derivfns}{2}
\derivitem{b.}
\catone{$\Gamma \yields A$}{}
\cattwo{$\Delta(B) \yields C$}{}
\reduce{1}{\lone}{\rtwo}{\bsl L}{$\Delta(\Gamma, A\bsl B)
\yields C$}
\derivitem{}
\catone{$A, \Gamma \yields B$}{}
\reduce{1}{\lone}{\rone}{\bsl R}{$\Gamma \yields A\bsl B$}
\end{derivfns}
\begin{derivfns}{2}
\derivitem{c.}
\catone{$\Gamma \yields A$}{}
\cattwo{$\Delta(B) \yields C$}{}
\reduce{1}{\lone}{\rtwo}{/L}{$\Delta(B/A, \Gamma) \yields C$}
\derivitem{}
\catone{$\Gamma, A \yields B$}{}
\reduce{1}{\lone}{\rone}{/R}{$\Gamma \yields B/A$}
\end{derivfns}
\begin{derivfns}{2}
\derivitem{d.}
\catone{$\Gamma(A, B) \yields C$}{}
\reduce{1}{\lone}{\rone}{$\product$L}{$\Gamma(A\product B) \yields C$}
\derivitem{}
\catone{$\Gamma\ \yields A$}{}
\cattwo{$\Delta \yields B$}{}
\reduce{1}{\lone}{\rtwo}{$\product$R}{$\Gamma, \Delta \yields A \product B$}
\end{derivfns}
Lambek showed Cut-elimination for both calculi, i.e.\ every
theorem has a Cut-free proof. Of the remaining rules each
instance of premises has exactly one connective occurrence
less than the corresponding conclusion so Cut-elimination
shows decidability through finite space Cut-free sequent
proof search from conclusions to premises.
Lifting is derivable in {\bf NL}
as follows:\\
\begin{derivfns}{4}
\derivnum
\catone{A\yields A}{}
\cattwo{B\yields B}{}
\reduce{1}{\lone}{\rtwo}{\bsl L}{[A, A\bsl B] \yields B}
\reduce{3}{\lone}{\rtwo}{/R}{A\yields B/(A\bsl B)}
\end{derivfns}
It is also derivable in {\bf L}; indeed all {\bf NL}
derivations are converted to {\bf L} derivations
by simply erasing the brackets. But {\bf L}-derivable
composition depends
essentially on associativity and is not {\bf NL}-derivable:\\
\begin{derivfns}{6}
\derivnum
\axiomone{2}{A\yields A}{}
\cattwo{B\yields B}{}
\catthree{C\yields C}{}
\reduce{1}{\ltwo}{\rthree}{\bsl L}{B, B\bsl C\yields C}
\reduce{3}{\lone}{\rthree}{\bsl L}{A, A\bsl B, B\bsl C\yields C}
\reduce{5}{\lone}{\rthree}{\bsl R}{A\bsl B, B\bsl C \yields A\bsl C}
\end{derivfns}
Even amongst the Cut-free proofs however there is still semantic
equivalence under the Curry-Howard rendering (van Benthem,
1983; see Morrill, 1994b)
and in this respect redundancy in parsing as exhaustive
proof search since
distinct lines of inference converge
on common subproblems.
This derivational equivalence (or: \scare{spurious ambiguity})
betrays the permutability of certain rule applications. Thus
two left rules may be permutable: N/CN, CN, N\bsl S\yields S
can be proved by choosing to work on either connective first.
And left and right rules are permutable: N/CN, CN\yields S/(N\bsl S))
may be proved by applying a left rule first, or a right rule,
(and the latter step then further admits the two
options of the first example). Such non-determinism
is not significant semantically: the variants have the same
readings;
the non-determinism in partitioning
by the binary left rules in \AL{}
{\em is} semantically significant, but still a source
of inefficiency in its backward chaining \scare{generate-and-test}
incarnation.
Another source of derivational equivalence is that a
complex id axiom instance such as N\bsl S\yields N\bsl S
can be proved either by a direct matching against the
axiom scheme, or by two rule applications. This is
easily solved by restricting id to atomic formulas. More
problematic are the
permutability of rule applications, the non-determinism
of rules requiring splitting of configurations in {\bf L},
and the need in {\bf NL} to hypothesise configuration
structure a priori (such hierarchical structure is not
given by the input to the parsing problem). It seems
that only the first of these difficulties can be overcome
from a Gentzen sequent perspective.


The situation regarding equivalence and rule ordering is solved,
at least for ${\bf L}{-}\{\product{\rm L}\}$, by sequent
proof normalisation
(K\"{o}nig, 1989;
Hepple, 1990; Hendriks, 1993):\\
\begin{derivfns}{2}
\derivnum
\derivitem{a.}
\catone{$\fbox{$A$} \yields A$}{}
\derivitem{id$^*$}
\catone{$\Gamma_1, \fbox{$A$}, \Gamma_2 \yields B$}{}
\reduce{1}{\lone}{\rone}{P$^*$}{$\Gamma_1, A, \Gamma_2 \yields
\fbox{$B$}$}
\end{derivfns}
\begin{derivfns}{2}
\derivitem{b.}
\catone{$\Gamma \yields \fbox{$A$}$}{}
\cattwo{$\Delta(\fbox{$B$}) \yields C$}{}
\reduce{1}{\lone}{\rtwo}{\bsl L$^*$}{$\Delta(\Gamma, \fbox{$A\bsl B$})
\yields C$}
\derivitem{}
\catone{$A, \Gamma \yields \fbox{$B$}$}{}
\reduce{1}{\lone}{\rone}{\bsl R}{$\Gamma \yields \fbox{$A\bsl B$}$}
\end{derivfns}
\begin{derivfns}{2}
\derivitem{c.}
\catone{$\Gamma \yields \fbox{$A$}$}{}
\cattwo{$\Delta(\fbox{$B$}) \yields C$}{}
\reduce{1}{\lone}{\rtwo}{/L$^*$}{$\Delta(\fbox{$B/A$}, \Gamma) \yields C$}
\derivitem{}
\catone{$\Gamma, A \yields \fbox{$B$}$}{}
\reduce{1}{\lone}{\rone}{/R}{$\Gamma \yields \fbox{$B/A$}$}
\end{derivfns}
This involves
firstly ordering right rules before
left rules reading from endsequent to axiom leaves
(so left rules only apply to sequents
with atomic succedents; this effects uniform
proof; see Miller et al., 1991), and
secondly
further demanding successive unfolding of the same
configuration type (\scare{focusing}).
In the *-ed rules the succedent is atomic. A necessary
condition for success is that an antecedent type is only
selected by P$^*$ if it yields the succedent atom as its
eventual range.
Let us refer to
(\prevex) as \fbox{\AL}. \fbox{\AL} is free of spurious ambiguity,
and $\vdash_{\mbox{\scriptsize \bf L}}\Gamma\yields A$ iff
$\vdash_{\fbox{\scriptsize \bf L}}\Gamma\yields \fbox{$A$}$.
The focusing strategy
breaks down for \product L: (VP/PP)/N, N\product PP\yields VP
requires switching between configuration types. It happens
that left occurrences of product are not motivated in
grammar, but more critically sequent proof normalisation
leaves the non-determinism of
partitioning, and offers no general method for multimodal
extensions which may have complex and interacting structural
properties. To eliminate the splitting
problem we need some kind of representation
of configurations such that the domain of functors need not
be hypothesised and then checked, but rather discovered
by constraint propagation. Such is the character of our
treatment, whereby partitioning is explored by unification
in the term structure of higher-order linear logic programming, to
which we now turn.
By way of orientation we review the (propositional)
features of clausal programming.

The first order case, naturally, corresponds to Prolog. Let
us assume a set ${\cal ATOM}$ of atomic
formulas, 0-ary, 1-ary, etc., formula constructors
$\{\cdot\wedge\ldots\wedge\cdot\}_{n\in\{0, 1, \ldots\}}$
and a binary (infix) formula constructor
$\leftarrow$. A sequent comprises an agenda formula $A$
and a database $\Gamma$ which is a bag of program clauses
$\{B_1, \ldots, B_n\}_m, n\ge 0$ (subscript $m$ for
multiset); we write $\Gamma\yields A$.
In BNF, the set of agendas corresponding to the
nonterminal ${\cal AGENDA}$ and the set of
program clauses corresponding to the nonterminal
${\cal PCLS}$ are defined by:\\
\begin{derivfns}{1}
\derivnum
\catone{\hspace*{-0.1in}$\begin{array}[t]{l}
{\cal AGENDA} ::= {\cal GOAL}\wedge\ldots\wedge{\cal GOAL}\\
{\cal PCLS} ::= {\cal ATOM}\leftarrow{\cal AGENDA}
\end{array}$}{}
\end{derivfns}
For first order programming the set ${\cal GOAL}$ of
goals is defined by:\\
\begin{derivfns}{0}
\derivnum
\catone{$
{\cal GOAL} ::= {\cal ATOM}
$}{}
\end{derivfns}
Then execution is guided by the following rules.\\
\begin{derivfns}{0}
\derivnum
\axiomone{0}{$\Gamma, A\yields A$}{}
\derivitem{ax}
\end{derivfns}
I.e.\ the unit agenda is a consequence of any database
containing its atomic clause.\\
\begin{derivfns}{3}
\derivnum
\axiomone{0.8}{$\Gamma, A\leftarrow B_1\wedge\ldots\wedge B_n\yields
\begin{array}{l}B_1\wedge\ldots\wedge B_n\wedge\\
C_1\wedge\ldots\wedge C_m
\end{array}$}{$\Gamma,
A\leftarrow B_1\wedge\ldots\wedge B_n\yields
A\wedge C_1\wedge\ldots\wedge C_m$}
\reduce{2}{\lone}{\rone}{RES}{$\Gamma,
A\leftarrow B_1\wedge\ldots\wedge B_n\yields
A\wedge C_1\wedge\ldots\wedge C_m$}
\end{derivfns}
I.e.\ we can resolve the first goal on the agenda with the head
of a program clause and then continue with the program as
before and a new agenda given by prefixing the program clause
subagenda to the rest of the original agenda (depth-first
search).

For the higher-order case agendas and program clauses are defined
as above, but the notion of ${\cal GOAL}$ on which they depend
is generalised to include implications:\\
\begin{derivfns}{0}
\derivnum
\catone{$
{\cal GOAL} ::= {\cal ATOM}\ |\ {\cal GOAL}\leftarrow{\cal PCLS}
$}{}
\end{derivfns}
And a \scare{deduction theorem} rule of inference is added:\\
\begin{derivfns}{2}
\derivnum
\catone{$\Gamma, B\yields A$}{}
\cattwo{$\Gamma\yields C_1\wedge\ldots\wedge
C_m$}{}
\reduce{1}{\lone}{\rtwo}{DT}{$\Gamma\yields (A\leftarrow B)\wedge
C_1\wedge\ldots\wedge
C_m$}
\end{derivfns}
I.e.\ we solve a higher-order goal first on the agenda by adding
its precondition to the database and trying to prove
its postcondition.

In linear logic programming
the rules become
resource conscious; in this context we write \mconj{} for
the conjunction and \lingate{} for the implication:\\
\begin{derivfns}{0}
\derivnum
\axiomone{0}{$A \yields A$}{}
\derivitem{ax}
\end{derivfns}
I.e.\ an atomic agenda is a consequence of its unit database:
all program clauses must be \scare{used up} by the resolution
rule:\\
\begin{derivfns}{2}
\derivnum
\catone{$\Gamma\yields
B_1\mconj\ldots\mconj B_n\mconj C_1\mconj\ldots\mconj
C_m$}{$\Gamma,
A\lingate B_1\mconj\ldots\mconj B_n\yields
A\mconj C_1\mconj\ldots\mconj C_m$}
\reduce{1}{\lone}{\rone}{RES}{$\Gamma,
A\lingate B_1\mconj\ldots\mconj B_n\yields
A\mconj C_1\mconj\ldots\mconj C_m$}
\end{derivfns}
I.e.\ a program clause disappears from the database once it
is resolved upon: each is used exactly once. The deduction theorem
rule for higher-order clauses also becomes sensitised to the
employment of antecedent contexts:\\
\begin{derivfns}{2}
\derivnum
\catone{$\Gamma, B\yields A$}{}
\cattwo{$\Delta\yields C_1\mconj\ldots\mconj
C_m$}{}
\reduce{1}{\lone}{\rtwo}{DT}{$\Gamma, \Delta\yields (A\lingate B)\mconj
C_1\mconj\ldots\mconj
C_m$}
\end{derivfns}


We shall motivate compilation into linear clauses directly
from simple algebraic models for the calculi. In the case of
{\bf L} we have first interpretation in semigroups
$\langle L, \add\rangle$ (i.e.\ sets $L$ closed under associative
binary operations \add; intuitively: strings
under concatenation). Relative to a model
each type $A$ has an interpretation as a subset $D(A)$
of $L$.
Given that primitive types are interpreted as some such
subsets, complex
types receive their denotations by {\em residuation} as follows
(cf.\ e.g.\ Lambek, 1988):\\
\begin{derivfns}{2}
\derivnum
\catone{\hspace*{-0.08in}$
\begin{array}[t]{lcl}
D(A\product B) & = & \{s_1\add s_2| s_1 \in D(A) \wedge s_2 \in D(B)\}\\
D(A\bsl B) & = & \{s| \forall s' \in D(A), s'\add s \in D(B)\}\\
D(B/A) & = & \{s| \forall s' \in D(A), s\add s' \in D(B)\}
\end{array}
$}{}
\end{derivfns}
For the non-associative calculus we drop the condition of
associativity and interpret in arbitrary groupoids (intuitively:
trees under adjunction\footnote{Though \NL{} with product
is incomplete with respect to finite trees as opposed to
groupoids in general.}).

Categorial type assignment statements comprise
a term $\alpha$ and a type $A$;
we write $\alpha\ass A$. Given a set of lexical assignments,
a phrasal assignment is projected if and only if in every
model satisfying the lexical assignments the phrasal
assignment is also satisfied. A categorial sequent
has a translation given by $|\cdot|$
into a linear sequent of type assignments which can be
safely read as predications. For {\bf L} we have the following
({\bf NL} preserves input antecedent configuration in output
succedent term structure):\\
\begin{derivfns}{1}
\derivnum
\catone{\hspace*{-0.08in}
$\begin{array}[t]{l}
|B_0, \ldots, B_n\yields A|\ =\\
\pcnst{k}_0\ass B_0^+, \ldots,
\pcnst{k}_n\ass B_n^+\yields
\pcnst{k}_0\add\ldots\add\pcnst{k}_n\ass A^-
\end{array}
$}{}
\end{derivfns}
Categorial type assignment
statements are translated into linear logic according
to the interpretation of types.
The polar translation functions are identity functions
on atomic assignments; on complex category
predicates they are defined mutually
as follows (for related unfolding,
but for proof nets, see Roorda, 1991; Moortgat, 1992;
Hendriks, 1993; and Oehrle, 1994); $\overline{p}$ indicates the polarity
complementary to $p$:\\
\begin{derivfns}{2}
\derivnum
\catone{$\alpha\add\gamma\ass B^p$}{}
\cattwo{$\lingate$}{}
\catthree{$\alpha\ass A^{\overline{p}}$}{}
\reduce{1}{\lone}{\rthree}{\begin{tabular}{l}$\alpha$ new variable/\\
constant as
$p\ {+}/{-}$
\end{tabular}}{$\gamma\ass A\bsl B^p$}
\end{deriv}
\begin{deriv}{2}
\catone{$\gamma\add\alpha\ass B^p$}{}
\cattwo{$\lingate$}{}
\catthree{$\alpha\ass A^{\overline{p}}$}{}
\reduce{1}{\lone}{\rthree}{\begin{tabular}{l}$\alpha$ new variable/\\
constant as
$p\ {+}/{-}$\end{tabular}}{$\gamma\ass B/A^p$}
\end{derivfns}
The unfolding transformations have the same general
form for the positive
(configurat\-ion/database) and negative (succedent/agenda)
occurrences; the polarity is used to indicate whether
new symbols introduced for quantified variables in
the interpretation clauses
are
metavariables (in italics) or Skolem constants
(in boldface); we shall see examples shortly.
The program clauses and agenda are read directly off the
unfoldings, with the only manipulation being a flattening
of positive implications into uncurried form:\\
\begin{derivfns}{1}
\derivnum
\catone{\hspace*{-0.08in}$\begin{array}[t]{l}
((X^+\lingate Y_1^-)\lingate \ldots)\lingate Y_n^-
\ \rhd\\
X^+\lingate Y_1^-\mconj\ldots\mconj Y_n^-
\end{array}$}{}
\end{derivfns}
(This means that matching against the head of a clause
and assembly of subgoals does not require any recursion
or restructuring at runtime.)
We shall also allow unit program clauses $X\lingate$ to
be abbreviated $X$.


Starting from the initial database and agenda, a proof will be
represented as a list of agendas, avoiding the context repetition
of sequent proofs by indicating where the resolution rule
retracts from the database (superscript
coindexed overline), and where the
deduction theorem rule adds to it (subscript coindexation):\\
\begin{derivfns}{3.5}
\derivnum
\catone{\hspace{-0.12in}
\begin{tabular}[t]{lll}
database & $\Gamma, \overline{A\lingate B_1\mconj\ldots\mconj
B_n}^{i}$\\
agenda\\
$i.$ & $A\mconj C_1\mconj\ldots\mconj C_m$ & RES\\
$i{+}1.$ & $B_1\mconj\ldots\mconj
B_n\mconj C_1\mconj\ldots\mconj C_m$
\end{tabular}}{}
\end{derivfns}
\begin{derivfns}{3}
\derivnum
\catone{\hspace*{-0.12in}
\begin{tabular}[t]{lll}
database & $\Gamma, {B}_i$\\
agenda\\
$i.$ & $(A\lingate B)\mconj C_1\mconj\ldots\mconj C_m$ & DT\\
$i{+}1.$ & $A\mconj C_1\mconj\ldots\mconj C_m$
\end{tabular}}{}
\end{derivfns}
The sharing of a Skolem constant between $A$ and $B$ in
(\prevex) ensures that $B$ can and must be used to prove
$A$ so that a mechanism for the lazy splitting of contexts
is effected.
The termination condition is met by a unit agenda with its unit
database.


By way of illustration for \AL{} consider
composition given the sequent translation (\nextex).\\
\begin{derivfns}{1}
\derivnum
\catone{\hspace*{-0.1in}\begin{tabular}[t]{l}
\vsl A\bsl B, B\bsl C\yields A\bsl C\vsl =\\
\pcnst{k}\ass A\bsl B$^+$, \pcnst{l}\ass B\bsl C$^+$\yields
\pcnst{k\add l}\ass A\bsl C$^-$
\end{tabular}}{}
\end{derivfns}
The assignments are unfolded thus:\\
\begin{plaindisp}
\begin{derivfns}{2}
\derivnum
\catone{$a\add\pcnst{k}$\ass B}{}
\cattwo{\lingate}{}
\catthree{$a$\ass A}{}
\reduce{1}{\lone}{\rthree}{}{\pcnst{k}\ass A\bsl B$^+$}
\derivitem{}
\catone{$b\add\pcnst{l}$\ass C}{}
\cattwo{\lingate}{}
\catthree{$b$\ass B}{}
\reduce{1}{\lone}{\rthree}{}{\pcnst{l}\ass B\bsl C$^+$}
\end{derivfns}
\begin{derivfns}{2}
\catone{\pcnst{m}\add(\pcnst{k}\add\pcnst{l})\ass C}{}
\cattwo{\lingate}{}
\catthree{\pcnst{m}\ass A}{}
\reduce{1}{\lone}{\rthree}{}{\pcnst{k}\add\pcnst{l}\ass A\bsl C$^-$}
\end{derivfns}
\end{plaindisp}
Then the proof runs as follows.\\
\begin{derivfns}{8.5}
\derivnum
\catone{\hspace*{-0.1in}
\begin{tabular}[t]{lll}
database &
\begin{minipage}[t]{1.25in}
$\overline{a\add\pcnst{k}\ass {\rm B}\lingate a\ass {\rm A}}^3$,\\
$\overline{b\add\pcnst{l}\ass {\rm C}\lingate b\ass {\rm B}}^2$,\\
$\overline{\pcnst{m}\ass {\rm A}_1}^4$
\end{minipage}\\
agenda\\
1. &
\pcnst{m}\add(\pcnst{k}\add\pcnst{l})\ass C\lingate\pcnst{m}\ass A
 & DT\\
2. & \pcnst{m}\add(\pcnst{k}\add\pcnst{l})\ass C
& RES $b=\pcnst{m}\add\pcnst{k}$\\
3. & \pcnst{m}\add\pcnst{k}\ass B & RES $a=\pcnst{m}$\\
4. & \pcnst{m}\ass A & RES
\end{tabular}}{}
\end{derivfns}
The unification at line~2 relies on associativity.
Note that unifications are all one-way, but even one-way
associative (=string) unification
has expensive worst cases.


For \NL{} the term labelling
provides a clausal implementation
with unification being non-associative. Consider lifting:\\
\begin{derivfns}{0}
\derivnum
\catone{\vsl A\yields B/(A\bsl B)\vsl\ =\
\pcnst{k}\ass A\yields \pcnst{k}\ass B/(A\bsl B)}{}
\end{derivfns}
\begin{derivfns}{4}
\derivnum
\axiomtwo{2}{\pcnst{k}\add\pcnst{l}\ass B}{}
\axiomthree{2}{\lingate}{}
\catthree{$a\add\pcnst{l}$\ass B}{}
\catfour{\lingate}{}
\catfive{$a$\ass A}{}
\reduce{1}{\lthree}{\rfive}{}{\pcnst{l}\ass A\bsl B$^+$}
\reduce{3}{\ltwo}{\rfive}{}{\pcnst{k}\ass B/(A\bsl B)$^-$}
\end{derivfns}
The proof is as follows.\\
\begin{derivfns}{6}
\derivnum
\catone{\hspace*{-0.1in}
\begin{tabular}[t]{lll}
database &
\begin{minipage}[t]{1.5in}
$
\overline{\pcnst{k}\ass A}^3,\\
\overline{a\add \pcnst{l}\ass {\rm B}\lingate a\ass A_1}^2
$
\end{minipage}
\\
agenda\\
1. &
$\pcnst{k}\add\pcnst{l}\ass {\rm B}\lingate
(a\add \pcnst{l}\ass {\rm B}\lingate a\ass A)$ & DT\\
2. & \pcnst{k}\add\pcnst{l}\ass {\rm B}
& RES $a=\pcnst{k}$\\
3. & \pcnst{k}\ass A & RES
\end{tabular}}{}
\end{derivfns}
The simple
one-way term unification is very fast but it is
unnatural from the point of view of parsing that,
as for the sequent approach, a hierarchical
binary structure on the input string needs to be
posited before inference begins, and exhaustive search would
require all possibilities to be tried. Later we shall
see how hierarchical structure can be discovered rather than
conjectured by factoring out horizontal structure.

Let us note here the relation to \fbox{\AL}. \fbox{\AL} applies
(working back from the target sequent)
right rules before left rules. Here, when a higher-order goal
is found on the agenda its precondition is added to the database
by DT.
This precedes applications of the RES rule (hence the uniformity
character) which corresponds to
the left sequent inferences. It applies when the agenda goal
is atomic and picks out antecedent types which yields
that atom (cf.\ the eventual range condition of
\fbox{\AL}). The focusing character is embodied by creating
in one step the objective of seeking all the arguments
of an uncurried functor.


By way of further example consider the following
in {\bf L}, with
terms
and types as indicated.\\
\begin{derivfns}{0}
\derivnum
\catone{(a book from which) the references are missing}{}
\end{derivfns}
\begin{derivfns}{4}
\derivnum
\wordone{the references}{\pcnst{r}\ass N}{}
\wordtwo{are missing}{\pcnst{m}\ass((S/(N\bsl S))\bsl S)/PP}{}
\nolinereduce{4}{\ltwo}{\rtwo}{\yields\pcnst{r}\add\pcnst{m}\ass S/PP}
\end{derivfns}
We have compilation for \lingform{are missing}
as in Figure~\ref{bigex1} yielding
(\nextex).\\
\begin{figure*}
\begin{derivfns}{6}
\derivitem{\tab\tab}
\axiomone{3}{$b\add(\pcnst{m}\add a)$\ass S}{}
\axiomtwo{3}{\lingate}{}
\axiomthree{1}{$b\add\pcnst{k}$\ass S}{}
\axiomfour{1}{\lingate}{}
\axiomfive{-1}{$c\add\pcnst{k}$\ass S}{}
\axiomsix{-1}{\lingate}{}
\axiomseven{-1}{$c$\ass N}{}
\axiomeight{5}{\lingate}{}
\axiomnine{5}{$a$\ass PP}{}
\reduce{0}{\lfive}{\rseven}{}{\pcnst{k}\ass N\bsl S$^+$}
\reduce{2}{\lthree}{\rseven}{}{$b$\ass S/(N\bsl S)$^-$}
\reduce{4}{\lone}{\rseven}{}{$\pcnst{m}\add a$\ass
(S/(N\bsl S))\bsl S$^+$}
\reduce{6}{\lone}{\rnine}{}{\pcnst{m}\ass((S/(N\bsl S))\bsl S)/PP$^+$}
\end{derivfns}
\caption{Groupoid compilation of the assignment to \lingform{are missing}}
\label{bigex1}
\end{figure*}
\begin{derivfns}{1}
\derivnum
\catone{\hspace*{-0.1in}\begin{tabular}[t]{l}
${\rhd}$\\$b\add(\pcnst{m}\add a)$\ass S\lingate($b\add\pcnst{k}$\ass
S\lingate($c\add\pcnst{k}$\ass S\lingate$c$\ass N))\mconj
$a$\ass PP
\end{tabular}}{}
\end{derivfns}
And the succedent unfolds as follows:\\
\begin{derivfns}{4}
\derivnum
\catone{(\pcnst{r}\add\pcnst{m})\add\pcnst{l}\ass S}{}
\cattwo{\lingate}{}
\catthree{\pcnst{l}\ass PP}{}
\reduce{1}{\lone}{\rthree}{}{\pcnst{r}\add\pcnst{m}\ass S/PP$^-$}
\nolinereduce{4}{\lone}{\rthree}{$\rhd\
$(\pcnst{r}\add\pcnst{m})\add\pcnst{l}\ass
S\lingate\pcnst{l}\ass PP}
\end{derivfns}
Derivation is as in figure~\ref{bigex2}.
\begin{figure*}
\begin{center}
\footnotesize
\begin{tabular}[t]{lll}
database &
\begin{minipage}[t]{2.5in}
$\overline{\mbox{\pcnst{r}\ass N}}^5$,\\
$\overline{\mbox{$b\add(\pcnst{m}\add a)$\ass
S\lingate($b\add\pcnst{k}$\ass
S\lingate($c\add\pcnst{k}$\ass S\lingate$c$\ass N))\mconj
$a$\ass PP
}}^2$,\\
$\overline{{\mbox{\pcnst{l}\ass PP}}_1}^6$,\\
$\overline{{\mbox{$c\add\pcnst{k}$\ass S\lingate
$c$\ass N}}_3}^4$,
\end{minipage}\\
agenda\\
1. &
(\pcnst{r}\add\pcnst{m})\add\pcnst{l}\ass
S\lingate\pcnst{l}\ass PP & DT\\
2. & (\pcnst{r}\add\pcnst{m})\add\pcnst{l}\ass
S & RES $b$=\pcnst{r}, $a$=\pcnst{l} \\
3. & (\pcnst{r}\add\pcnst{k}\ass
S\lingate($c\add\pcnst{k}$\ass S\lingate$c$\ass N))\mconj
\pcnst{l}\ass PP & DT\\
4. & \pcnst{r}\add\pcnst{k}\ass
S\mconj
\pcnst{l}\ass PP & RES $c$=\pcnst{r}\\
5. & \pcnst{r}\ass N\mconj
\pcnst{l}\ass PP & RES\\
6. & \pcnst{l}\ass PP & RES
\end{tabular}
\end{center}
\caption{Groupoid execution for \lingform{the references are missing}}
\label{bigex2}
\end{figure*}
The unification at line 2 relies on
associativity and as always
atomic goals on the agenda
are ground. But in general we have to
try subproofs for different
unifiers, that is, we effectively still have to guess
partitioning for left rules. We
shall see that this is not necessary,
and that associative unification can be avoided.


There is a further problem which will be solved in the same
move. Unfolding of left products would
create two positive subformulas and thus fall outside the scope
of Horn clause programming. However, the term-labelled
implementation as it has been given also fails for right
products:\\
\begin{derivfns}{2}
\derivnum
\catone{$\alpha\ass A^-$}{}
\cattwo{\mconj}{}
\catthree{$\beta\ass B^-$}{}
\reduce{1}{\lone}{\rthree}{$\gamma=\alpha\add\beta$?}{$\gamma\ass
A\product B^-$}
\end{derivfns}
The problem is that $\alpha$ and $\beta$ are not deterministically
given by $\gamma$ at the \scare{compile time} of unfolding. The
best we could manage seems to be to try different partitionings
of $\gamma$ at execution time; but even if this could work
it would still amount to trying different partitionings for
\product{}R as in the sequent calculus: a source of non-determinism
we seek to reduce. This limitation combines with the other
difficulties with groupoid labelling
of worst case of (even) one-way associative unification
for {\bf L}, and
the need for a priori hypothesis of non-associative structure
for {\bf NL}.


The method of solution resides in looking at an alternative model:
the associative calculus has
relational algebraic models (van Benthem, 1991) which
interpret types as relations on some set $V$, i.e.\ as sets of
ordered pairs. Given denotations for primitive types, those
of compound types are fixed as subsets of $V\times V$ by:\\
\begin{derivfns}{6}
\derivnum
\catone{\hspace*{-0.1in}
$
\begin{array}[t]{lcl}
D(A\bsl B) & = & \{\langle v_2, v_3\rangle|\forall\langle v_1,
v_2\rangle\in D(A),\\
 & & \hfill \langle v_1, v_3\rangle\in D(B)\}\\
D(B/A) & = & \{\langle v_1, v_2\rangle|\forall\langle v_2,
v_3\rangle\in D(A),\\
 & & \hfill \langle v_1, v_3\rangle\in D(B)\}\\
D(A\product B) & = & \{\langle v_1, v_3\rangle|
\exists v_2, \langle v_1, v_2\rangle\in
D(A)\ \&\ \\
 & & \hfill \langle v_2, v_3\rangle\in
D(B)\}
\end{array}
$}{}
\end{derivfns}
Points in $V$ intuitively corresponds to string positions
(as in definite clause grammars, and charts)
and ordered pairs to the vertices of substrings pertaining to
the categories to which they are assigned.
This induces unfolding as follows:\\
\begin{derivfns}{2}
\derivnum
\catone{$i\lnk k\ass B^p$}{}
\cattwo{$\lingate$}{}
\catthree{$i\lnk j\ass A^{\overline{p}}$}{}
\reduce{1}{\lone}{\rthree}{\begin{tabular}{l}$i$ new variable/\\constant as
$p\ {+}/{-}$\end{tabular}}{$j\lnk k\ass A\bsl B^p$}
\end{derivfns}
\begin{derivfns}{2}
\catone{$i\lnk k\ass B^p$}{}
\cattwo{$\lingate$}{}
\catthree{$j\lnk k\ass A^{\overline{p}}$}{}
\reduce{1}{\lone}{\rthree}{\begin{tabular}{l}$k$ new variable/\\constant as
$p\ {+}/{-}$\end{tabular}}{$i\lnk j\ass B/A^p$}
\end{derivfns}
Furthermore right product (though still not non-Horn
left product) unfolding can be expressed:\\
\begin{derivfns}{2}
\derivnum
\catone{$i\lnk j\ass A^-$}{}
\cattwo{\mconj}{}
\catthree{$j\lnk k\ass B^-$}{}
\reduce{1}{\lone}{\rthree}{$j$ new variable}{$i\lnk k\ass
A\product B^-$}
\end{derivfns}

Composition is now treated as follows. Assume sequent translation
thus:\\
\begin{derivfns}{1}
\derivnum
\catone{\hspace*{-0.1in}\begin{tabular}[t]{l}
\vsl A\bsl B, B\bsl C\yields A\bsl C\vsl\ =\\
0\lnk{1}\ass A\bsl B$^+$,
{1}\lnk{2}\ass B\bsl C$^+$\yields
{0}\lnk{2}\ass A\bsl C$^-$
\end{tabular}}{}
\end{derivfns}
The assignments are compiled as shown in (\nextex).
\begin{plaindisp}
\begin{derivfns}{2}
\derivnum
\catone{$i\lnk{1}$\ass B}{}
\cattwo{\lingate}{}
\catthree{$i\lnk 0$\ass A}{}
\reduce{1}{\lone}{\rthree}{}{0\lnk 1\ass A\bsl B$^+$}
\end{derivfns}
\begin{derivfns}{2}
\catone{$j\lnk 2$\ass C}{}
\cattwo{\lingate}{}
\catthree{$j\lnk{1}$\ass B}{}
\reduce{1}{\lone}{\rthree}{}{1\lnk 2\ass B\bsl C$^+$}
\catfour{3\lnk 2\ass C}{}
\catfive{\lingate}{}
\catsix{3\lnk 0\ass A}{}
\reduce{1}{\lfour}{\rsix}{}{0\lnk 2\ass A\bsl C$^-$}
\end{derivfns}
\end{plaindisp}
The proof is thus:\\
\begin{derivfns}{8.5}
\derivnum
\catone{\hspace*{-0.1in}
\begin{tabular}[t]{lll}
database &
\begin{minipage}[t]{1.25in}
$\overline{i\lnk 1\ass {\rm B}\lingate i\lnk 0\ass {\rm A}}^3$,\\
$\overline{j\lnk 2\ass {\rm C}\lingate j\lnk 1\ass {\rm B}}^2$,\\
$\overline{3\lnk 0\ass {\rm A}_1}^4$
\end{minipage}\\
agenda\\
1. &
3\lnk 2\ass C\lingate3\lnk 0\ass A
 & DT\\
2. & 3\lnk 2\ass C
& RES $j=3$\\
3. & 3\lnk 1\ass B & RES $i=3$\\
4. & 3\lnk 0\ass A & RES
\end{tabular}}{}
\end{derivfns}
In this way associative unification is avoided; indeed the
only matching is trivial unification between constants
and variables. So for \AL{} the relational
compilation allows partitioning by the binary rules to
be discovered by simple constraint propagation rather
than by the generate-and-test strategy of normalised
sequent proof.


Although the (one-way) term unification for groupoid
compilation of the non-associative calculus is very
fast we want to get round the fact that a hierarchical
binary structure on the input string needs to be
posited before inference begins. We can do this through
observation of the
following:
\begin{itemize}
\item
All non-associative theorems are associative theorems
(ignore brackets)
\item
Interpret non-associative operators in the product algebra
of \NL{} groupoid algebra and \AL{} relational
algebra, and perform labelled compilation accordingly
\item
Use the (efficient) relational labelling to check associative
validity
\item
Use the groupoid labelling to
check non-associative validity
and
compute the prosodic form induced
\end{itemize}
I.e.\ the endsequent succedent groupoid term can be left as
a variable and the groupoid unification performed
on the return trip from axiom leaves after associative
validity has been assured, as will be seen in
our final example. The groupoid unification will
now be one-way in the opposite direction.

The simultaneous compilation separates horizontal
structure (word order) represented by interval
segments, and horizontal-and-vertical
structure (linear and
hierarchical organisation) represented by groupoid
terms,
and uses the efficient segment labelling to compute
{\bf L}-validity, and then the term labelling both to check
the stricter {\bf NL}-validity, and to calculate the
hierarchical structure. In this way we use the fact that models
for {\bf NL} are given by intersection in
the product of relational
and groupoid models. Each type $A$ has an interpretation
$D(A)$ as a subset of $L\times V\times V$:
\begin{derivfns}{6}
\derivnum
\catone{\hspace*{-0.1in}
$
\begin{array}[t]{lcl}
D(A\bsl B) & = & \{\langle s, v_2, v_3\rangle|\forall\langle s', v_1,
v_2\rangle\in D(A),\\
 & & \hfill \langle s'\add s, v_1, v_3\rangle\in D(B)\}\\
D(B/A) & = & \{\langle s, v_1, v_2\rangle|\forall\langle s', v_2,
v_3\rangle\in D(A),\\
 & & \hfill \langle s\add s', v_1, v_3\rangle\in D(B)\}\\
D(A\product B) & = & \{\langle s_1\add s_2, v_1, v_3\rangle|
\exists v_2, \langle s_1, v_1, v_2\rangle\in
D(A)\\
 & & \hfill \ \&\ \langle s_2, v_2, v_3\rangle\in
D(B)\}
\end{array}$}{}
\end{derivfns}
Unfolding is thus:\\
\begin{plaindisp}
\begin{derivfns}{3}
\derivnum
\axiomone{1}{$\alpha\add\gamma\lnksm i\lnksm k\ass B^p$}{}
\axiomtwo{1}{$\lingate$}{}
\axiomthree{1}{$\alpha\lnksm i\lnksm j\ass A^{\overline{p}}$}{}
\reduce{2}{\lone}{\rthree}{\begin{tabular}{l}$\alpha, i$ new variables/\\
constants as
$p\ {+}/{-}$\end{tabular}}{$\gamma\lnksm j\lnksm k\ass A\bsl B^p$}
\end{derivfns}
\begin{derivfns}{2}
\catone{$\gamma\add\alpha\lnksm i\lnksm k\ass B^p$}{}
\cattwo{$\lingate$}{}
\axiomthree{0}{$\alpha\lnksm j\lnksm k\ass A^{\overline{p}}$}{}
\reduce{1}{\lone}{\rthree}{\begin{tabular}{l}$\alpha, k$ new
variables/\\
constants as
$p\ {+}/{-}$\end{tabular}}{$\gamma\lnksm i\lnksm j\ass B/A^p$}
\end{derivfns}
\begin{derivfns}{1}
\axiomone{0}{$\alpha\lnksm i\lnksm j\ass A^-$}{}
\axiomtwo{0}{\mconj}{}
\axiomthree{0}{$\beta\lnksm j\lnksm k\ass B^-$}{}
\reduce{1}{\lone}{\rthree}{$\alpha, \beta, j$ new variables}{$\alpha\add
\beta\lnksm i\lnk k\ass
A\product B^-$}
\end{derivfns}
\end{plaindisp}

By way of example consider the following:
\begin{plaindisp}
\begin{derivfns}{2}
\derivnum
\wordone{the references}{\pconst{r}\lnksm0\lnksm1\ass N}{}
\wordtwo{are missing}{\pconst{m}\lnksm1\lnksm2\ass((S/(N\bsl S))\bsl
S)/PP}{}
\end{derivfns}
\begin{derivfns}{2}
\wordthree{from this book}{\pconst{f}\lnksm2\lnksm3\ass PP}{}
\axiomfour{2}{\yields$d\lnksm 0\lnksm 3$\ass S}{}
\end{derivfns}
\end{plaindisp}
The unfolding compilation yielding (\nextex) for \lingform{are missing}
is given in Figure~\ref{bigex3}.\\
\begin{figure*}
\begin{derivfns}{6}
\derivitem{}
\axiomone{3}{$b\add(\pconst{m}\add a)\lnksm i\lnksm k_1$\ass S}{}
\axiomtwo{3}{\lingate}{}
\axiomthree{1}{$b\add\pconst{k}\lnksm i\lnksm4$\ass S}{}
\axiomfour{1}{\lingate}{}
\axiomfive{-1}{$c\add\pconst{k}\lnksm l\lnksm 4$\ass S}{}
\axiomsix{-1}{\lingate}{}
\axiomseven{-1}{$c\lnksm l\lnksm 1$\ass N}{}
\axiomeight{5}{\lingate}{}
\axiomnine{5}{$a\lnksm2\lnksm k_1$\ass PP}{}
\reduce{0}{\lfive}{\rseven}{}{\pconst{k}\lnksm 1\lnksm 4\ass N\bsl S$^+$}
\reduce{2}{\lthree}{\rseven}{}{$b\lnksm i\lnksm1$\ass S/(N\bsl S)$^-$}
\reduce{4}{\lone}{\rseven}{}{$\pconst{m}\add a\lnksm1\lnksm k_1$\ass
(S/(N\bsl S))\bsl S$^+$}
\reduce{6}{\lone}{\rnine}{}{\pconst{m}\lnksm1\lnksm2\ass((S/(N\bsl S))\bsl
S)/PP$^+$}
\end{derivfns}
\caption{Groupoid-relational compilation of the assignment to
\lingform{are
missing}}
\label{bigex3}
\end{figure*}
\begin{derivfns}{2}
\derivnum
\catone{\hspace*{-0.1in}\begin{tabular}[t]{l}${\rhd}$\\
$b\add(\pconst{m}\add a)\lnksm i\lnksm k_1$\ass
S\lingate\\
($b\add\pconst{k}\lnksm i\lnksm 4$\ass S
\lingate($c\add\pconst{k}\lnksm l\lnksm 4$\ass
S\lingate$c\lnksm l\lnksm 1$\ass N))\mconj
$a\lnksm 2\lnksm k_1$\ass PP
\end{tabular}}{}
\end{derivfns}
The derivation is given in Figure~\ref{bigex4}.
\begin{figure*}
\begin{center}
\footnotesize
\begin{tabular}[t]{lll}
database &
\begin{minipage}[t]{2.5in}
$\overline{\pconst{r}\lnksm0\lnksm1\ass\mbox{N}}^4$,\\
$\overline{b\add(\pconst{m}\add a)\lnksm i\lnksm k_1\ass
\mbox{S}\lingate(b\add\pconst{k}\lnksm i\lnksm 4\ass
\mbox{S}\lingate(c\add\pconst{k}\lnksm l\lnksm 4\ass
\mbox{S}\lingate c\lnksm l\lnksm 1\ass \mbox{N}))\mconj
a\lnksm 2\lnksm k_1\ass \mbox{PP}}^1$,\\
$\overline{{c\add\pconst{k}\lnksm l\lnksm 4\ass
\mbox{S}\lingate c\lnksm l\lnksm 1\ass \mbox{N}}_2}^3$,\\
$\overline{\pconst{f}\lnksm 2\lnksm 3\ass\mbox{PP}}^5$
\end{minipage}\\
agenda\\
1. & $d\lnksm 0\lnksm 3$\ass S & RES $d=b\add(\pconst{m}\add a)$\\
2. & $(b\add\pconst{k}\lnksm 0\lnksm 4$\ass
\mbox{S}\lingate($c\add\pconst{k}\lnksm l\lnksm 4$\ass
\mbox{S}\lingate $c\lnksm l\lnksm 1$\ass \mbox{N}))\mconj
$a\lnksm 2\lnksm 3$\ass \mbox{PP} & DT\\
3. & $b\add\pconst{k}\lnksm 0\lnksm 4$\ass S\mconj $a\lnksm2\lnksm3$\ass
PP & RES $b=c$\\
4. & $c\lnksm0\lnksm1$\ass N\mconj$a\lnksm 2\lnksm 3$\ass PP
& RES $c=\pconst{r}$\\
5. & $a\lnksm2\lnksm3$\ass PP & RES $a=\pconst{f}$
\end{tabular}
\end{center}
\caption{Groupoid-relational execution for \lingform{the references
are missing from this book}}
\label{bigex4}
\end{figure*}
Note how the term unification computing the hierarchical
structure can be carried out one-way in the reverse order
to the forward segment matchings:
\begin{derivfns}{1}
\derivnum
\catone{\hspace*{-0.08in}$\begin{array}[t]{l}
d = b\add(\pconst{m}\add a) =
c\add(\pconst{m}\add a) =
\pconst{r}\add(\pconst{m}\add a) =\\
\pconst{r}\add(\pconst{m}\add \pconst{f})
\end{array}
$}{}
\end{derivfns}
In the case of {\bf NL}-invalidity the term unification
would fail.


We mention finally multimodal generalisations.
In multimodal calculi families of
connectives $\{/_i, \bsl{}_i,\product_i\}_{i\in \{1, \ldots, n\}}$
are
each defined by residuation with respect to their adjunction
in a \scare{polygroupoid} $\langle L, \{\add_i\}_{i\in
\{1, \ldots, n\}}\rangle$ (Moortgat and Morrill, 1991):\\
\begin{derivfns}{2.5}
\derivnum
\catone{\hspace*{-0.1in}
$
\begin{array}[t]{lcl}
D(A\product_i B) & = & \{s_1\add_i s_2| s_1 \in D(A) \wedge s_2 \in D(B)\}\\
D(A\bsl_i B) & = & \{s| \forall s' \in D(A), s'\add_i s \in D(B)\}\\
D(B/_iA) & = & \{s| \forall s' \in D(A), s\add_i s' \in D(B)\}
\end{array}
$}{}
\end{derivfns}
Multimodal groupoid compilation for implications is
immediate:\\
\begin{plaindisp}
\begin{derivfns}{2}
\derivnum
\catone{$\alpha\add_i\gamma\ass B^p$}{}
\cattwo{$\lingate$}{}
\catthree{$\alpha\ass A^{\overline{p}}$}{}
\reduce{1}{\lone}{\rthree}{\begin{tabular}{l}$\alpha$ new variable/\\
constant as
$p\ {+}/{-}$\end{tabular}}{$\gamma\ass A\bsl_i B^p$}
\end{deriv}
\begin{deriv}{2}
\catone{$\gamma\add_i\alpha\ass B^p$}{}
\cattwo{$\lingate$}{}
\catthree{$\alpha\ass A^{\overline{p}}$}{}
\reduce{1}{\lone}{\rthree}{\begin{tabular}{l}$\alpha$ new variable/\\
constant as
$p\ {+}/{-}$\end{tabular}}{$\gamma\ass B/_iA^p$}
\end{derivfns}
\end{plaindisp}
This is entirely general. Any multimodal calculus can be implemented
this way provided we have a (one-way) unification algorithm
specialised according to the structural communication axioms.
For example Morrill (1993)
deals with multimodality for discontinuity
which involves varying internal structural properties
(associativity vs. non-associativity) as well as
\scare{split/wrap}
interaction between modes.
This is treated computationally in the current manner in Morrill (1994a)
which also considers head-oriented discontinuity
and unary
operators projecting bracketed string structure. In these
cases also simultaneous compilation including binary relational
labelling can provide additional advantages.


Labelled unfolding of categorial formulas has been invoked
in the references cited as a way of checking well-formedness
of proof nets for categorial calculi by unification
of labels on linked formulas. This offers improvements
over sequent formulations but raises alternative problems; for
example associative unification in general can have infinite
solutions and is undecidable.
Taking linear validity as the highest common factor of
sublinear categorial calculi
we have been able
to show a strategy based on resolution in which the flow
of information is such that one term in unification is always
ground. Furthermore binary relational labelling propagates
constraints in such a way that computation of unifiers
may be reduced to a subset of cases or
avoided altogether. Higher-order coding allows emission
of hypotheticals
to be postponed until they are germane. Simultaneous
compilation allows a factoring out of horizontal structure from
vertical structure within the sublinear space in such
a way that the partial information of word order
can drive computation of hierarchical structure
for the categorial parsing problem in the presence
of non-associativity. The treatments for the calculi
above and their multimodal generalisations have been
implemented in Prolog (Morrill, 1994a).

\section*{References}

\setlength{\parindent}{0in}
\setlength{\parskip}{0.035in}

\ent{van Benthem, Johan: 1983, `The Semantics of Variety in Categorial
Grammar', Report 83-29, Department of Mathematics, Simon Fraser
University, also in
Buszkowski,~W., W.~Marciszewski, and J.~van Benthem
(eds.): 1988, {\it Categorial Grammar}, Linguistic
\& Literary Studies in Eastern Europe Volume~25, John Benjamins,
Amsterdam, 37--55.}

\ent{van Benthem,~J.: 1991, {\it Language in Action:
Categories, Lambdas and Dynamic Logic\/}, Studies in
Logic and the Foundations of Mathematics Volume~130,
North-Holland, Amsterdam.}

\ent{Hendriks, Herman: 1993, {\it Studied Flexibility:
Categories and Types in Syntax and Semantics}, Ph.D
dissertation, Institute for Logic, Language and Computation,
Universiteit van Amsterdam.}

\ent{Hepple, Mark: 1990, {\it The Grammar and Processing of Order and
Dependency: A Categorial Approach}, Ph.D. dissertation,
University of Edinburgh.}


\ent{Hepple, Mark: 1995, `Mixing Modes of Linguistic Description
in Categorial Grammar', this volume.}

\ent{Hodas, J.: 1992, `Specifying Filler-Gap Dependency Parsers
in a Linear-Logic Programming Language', in {\it Proceedings
of the Joint International Conference and
Symposium on Logic Programming}, 622--636.}

\ent{Hodas, Joshua and Dale Miller: 1994, `Logic Programming
in a Fragment of Intuitionistic Linear Logic', to appear
in {\it Journal of Information and Computation}.}

\ent{K\"{o}nig, E.: 1989, `Parsing as natural deduction',
in {\it Proceedings of the Annual Meeting of the Association
for Computational Linguistics}, Vancouver.}

\ent{Lambek,~J.: 1958, `The mathematics of sentence structure',
{\it American Mathematical Monthly\/}~{\bf 65}, 154--170,
also in
Buszkowski,~W., W.~Marciszewski, and J.~van Benthem
(eds.): 1988, {\it Categorial Grammar}, Linguistic
\& Literary Studies in Eastern Europe Volume~25, John Benjamins,
Amsterdam, 153--172.}

\ent{Lambek,~J.: 1961, `On the calculus of syntactic types', in
R.~Jakobson (ed.)
{\it Structure of language and its mathematical aspects},
Proceedings of the Symposia in Applied Mathematics {\bf XII},
American Mathematical Society, 166--178.}

\ent{Lambek,~J.: 1988, `Categorial and Categorical Grammars',
in Richard~T. Oehrle, Emmon Bach, and Deidre Wheeler (eds.)
{\it Categorial Grammars and Natural Language Structures},
Studies in Linguistics and Philosophy Volume~32, D.~Reidel,
Dordrecht, 297--317.}

\ent{Miller,~D., G.~Nadathur, F.~Pfenning, and A.~Scedrov:
1991, `Uniform Proofs as a Foundation for Logic
Programming', {\it Annals of Pure and Applied Logic}
{\bf 51}, 125--157.}

\ent{Moortgat, Michael: 1988, {\it Categorial Investigations:
Logical and Linguistic Aspects of the Lambek Calculus},
Foris, Dordrecht.}

\ent{Moortgat, Michael: 1992, `Labelled Deductive Systems for
categorial theorem proving', OTS Working Paper
OTS--WP--CL--92--003, Rijksuniversiteit Utrecht, also
in {\it Proceedings of
the Eighth Amsterdam Colloquium}, Institute for Language,
Logic and Information, Universiteit van Amsterdam.}

\ent{Moortgat, Michael and Glyn Morrill: 1991, `Heads and Phrases:
Type Calculus for Dependency and Constituent Structure',
to appear in {\it Journal of Language, Logic, and
Information}.}

\ent{Moortgat, Michael and Dick Oehrle: 1994, `Adjacency,
dependency and order', in {\it Proceedings of the Ninth
Amsterdam Colloquium}, 447--466.}

\ent{Morrill, Glyn: 1993, {\it Discontinuity and Pied-Piping
in Categorial Grammar}, Report de Recerca LSI--93--18--R,
Departament
de Llenguatges i Sistemes Inform\`{a}tics,
Universitat Polit\`{e}cnica de Catalunya, to appear in
{\it Linguistics and Philosophy}.}

\ent{Morrill, Glyn: 1994a, `Higher-Order Linear Logic Programming
of Categorial Deduction', Report de Recerca LSI--94--42--R,
Departament
de Llenguatges i Sistemes Inform\`{a}tics,
Universitat Polit\`{e}cnica de Catalunya}

\ent{Morrill, Glyn: 1994b, {\it Type Logical Grammar: Categorial
Logic of Signs}, Kluwer Academic Publishers, Dordrecht.}


\ent{Oehrle, Dick: 1994, `Term labelled categorial type
systems', to appear in {\it Linguistics and Philosophy}.}

\ent{Pareschi, R.: 1989, {\it Type-driven Natural Language
Analysis}, Ph.D. thesis, University of Edinburgh.}

\ent{Pareschi, R. and D, Miller: 1990, `Extending Definite Clause
Grammars with Scoping Constructs', in D.H.D. Warren and P. Szeredi
(eds.) {\it 1990 International Conference in Logic
Programming}, MIT Press, 373--389.}

\ent{Roorda, Dirk: 1991, {\it Resource Logics: proof-theoretical
investigations}, Ph.D. dissertation, Universiteit van Amsterdam.}

\end{document}